\pdfoutput=1

\documentclass[11pt]{article}

\usepackage[]{acl}

\usepackage{times}
\usepackage{latexsym}

\usepackage[T1]{fontenc}

\usepackage[utf8]{inputenc}

\usepackage{microtype}

\usepackage{inconsolata}

\usepackage{adjustbox}
\usepackage{booktabs}
\usepackage{xltabular}
\usepackage{multirow}
\usepackage{svg}
\usepackage{subcaption}
\usepackage{caption}
\usepackage[nameinlink,capitalize,noabbrev]{cleveref}

\crefname{equation}{equation}{equations}   
\crefname{footnote}{footnote}{footnotes}   
\crefname{section}{\S}{\S\S}
\Crefname{section}{\S}{\S\S}    
\crefformat{section}{#2\S#1#3}  
\Crefformat{section}{#2\S#1#3}
\crefrangeformat{section}{\S\S#3#1#4--#5#2#6}
\Crefrangeformat{section}{\S\S#3#1#4--#5#2#6}
\crefformat{section}{#2\S#1#3}  
\crefrangeformat{section}{\S\S#3#1#4--#5#2#6}
\usepackage{fnpct}
\usepackage{hyperref}
\usepackage{xcolor}

\title{Since the Scientific Literature Is Multilingual, Our Models Should Be Too}

\author{Abteen Ebrahimi${ }^{1}$ \and Kenneth Church${ }^{2}$ \\
  ${ }^{1}$University of Colorado Boulder, ${ }^{2}$Northeastern University \\
  \texttt{abteen.ebrahimi@colorado.edu, k.church@northeastern.edu}}

\begin{document}
\maketitle
\begin{abstract}
English has long been assumed the \textit{lingua franca} of scientific research, and this notion is reflected in the natural language processing (NLP) research involving scientific document representation. In this position piece, we quantitatively show that the literature is largely multilingual and argue that current models and benchmarks should reflect this linguistic diversity. We provide evidence that text-based models fail to create meaningful representations for non-English papers and highlight the negative user-facing impacts of using English-only models non-discriminately across a multilingual domain. We end with suggestions for the NLP community on how to improve performance on non-English documents.
\end{abstract}

\section{Introduction}

There has been a noticeable shift in recent years away from the traditional English-centric focus of traditional natural language processing (NLP) research. Pretrained models, which drive many state of the art approaches, have moved from explicitly monolingual language coverage \cite{devlinBERTPretrainingDeep2019, radfordLanguageModelsAre2019} to strong multilingual support \cite{brownLanguageModelsAre2020, workshopBLOOM176BParameterOpenAccess2023}. Machine translation systems, multilingual by definition, are now aiming at the long tail of languages, to create usable translations for languages with minimal resources \cite{nllbteamNoLanguageLeft2022, bapnaBuildingMachineTranslation2022}. Major NLP conferences and workshops have begun to heavily promote research involving other languages\footnote{\href{https://www.2022.aclweb.org/post/acl-2022-theme-track-language-diversity-from-low-resource-to-endangered-languages}{ACL '22}, \href{https://2024.naacl.org/blog/NAACL-2024-Theme-Track-Languages-of-Latin-America/}{NAACL '24}}, particularly those which are low-resource or endangered\footnote{\href{https://turing.iimas.unam.mx/americasnlp/}{AmericasNLP}, \href{https://www.masakhane.io/}{Masakhane}}. This shift is well motivated; methods which are truly language agnostic and robust to varying amounts of data are technologically interesting, and from a human-centered perspective, we must strive to create language technologies which are more inclusive to speakers of all languages around the world \cite{elraInternationalConferenceLanguage2019, joshiStateFateLinguistic2020}. 
\begin{table}[t]
\small
\begin{tabularx}{\linewidth}{lX}
        \toprule
         ID & \href{https://www.semanticscholar.org/paper/d831655c4c5246fbafe447f7c36096a9ec084512}{84779794}\\
         Title & \textit{Relation between ants and aphids in a citrus orchard.} \\
         TL;DR & Five-year-old boy with Down's syndrome is diagnosed with encephalitis.\\
         \bottomrule

\end{tabularx}  
    \caption{    \label{tab:my_label}
    Example from Semantic Scholar \cite{Wade2022TheSS,Kinney2023TheSS} where attempts to standardize on English introduce confusions. The paper was published in Japanese, but the title and tl;dr fields 
    are in English. Additional examples can be seen in \cref{tab:real_world_tab}.
    }

\end{table}

\begin{figure*}[t]
    \centering
    \includegraphics[width=\linewidth]{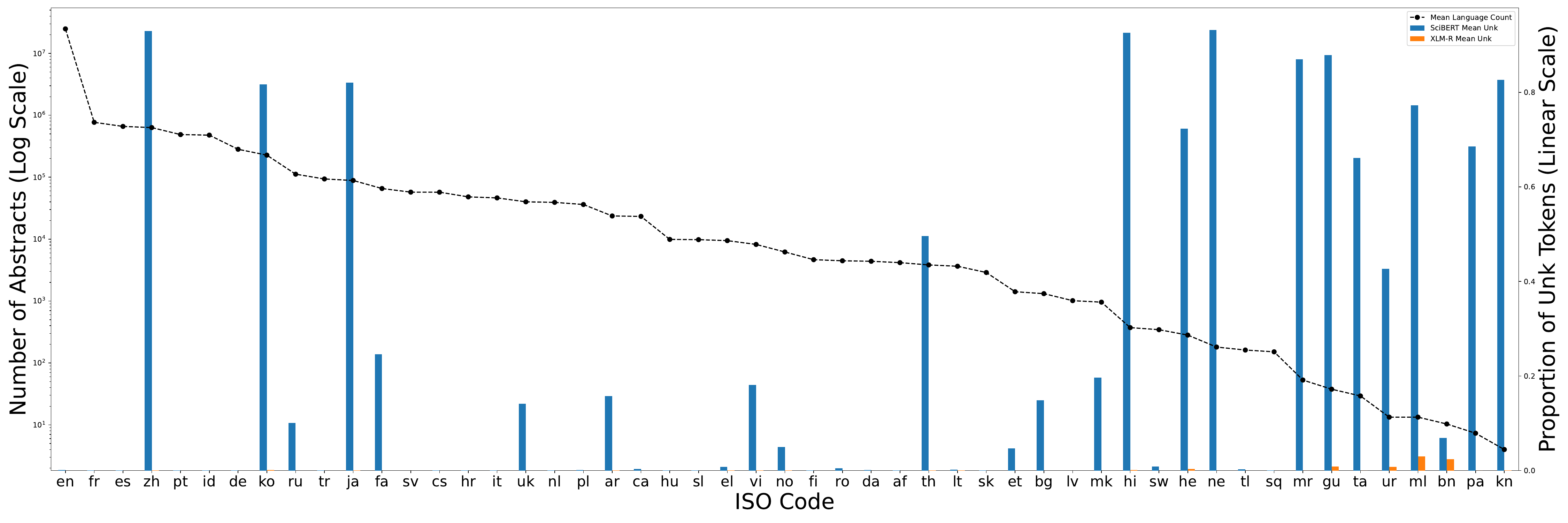}
    \caption{Main results. Black dotted line represents the total number of abstracts for each language. Bars represented the proportion of tokens which are unknown on average for that language, for a monolingual vs. multilingual model.}
    \label{fig:main_fig}
\end{figure*}

In this position piece, we highlight one area of NLP research which we believe stands to benefit from improving its multilingual support: scientific document representation. English has been historically considered the \textit{lingua franca} of scientific research, to the disadvantage of many non-English speaking communities \cite{hanauerLinguisticInjusticeWriting2019, marquezScienceCommunicationMultiple2020, ramirez-castanedaDisadvantagesPreparingPublishing2020,steigerwaldOvercomingLanguageBarriers2022}. This belief is also reflected in prior NLP research which involves scientific literature. By and large, most of the currently published models and datasets only support English. In this paper, we show that, while English makes up the dominant proportion of the literature, the remainder is incredibly multilingual. We further show that 
English-only models 
fail to create meaningful representations for many languages. Finally, we present the real world effects of the English-only models and discuss future steps to help better process documents in other languages.  

We note that this work is specific to the NLP community; the broader scope of issues and solutions surrounding English as the language of research are well-studied by other fields and far-reaching. We focus on the technical issues which arise from using English-only models, and consider the larger discussion out of scope for this work.

\section{Related Work}



\citet{beltagySciBERTPretrainedLanguage2019} present SciBERT, a BERT-style \cite{devlinBERTPretrainingDeep2019a} model trained on papers taken from Semantic Scholar \cite{Wade2022TheSS,Kinney2023TheSS}, largely from the computer science and biomedical domains. 
SciBERT is the base of document representation models such as Specter \cite{cohanSPECTERDocumentlevelRepresentation2020}, through a post-pretraining alignment using the citation graph. 
Evaluation benchmarks for document representation include SciDocs \cite{cohanSPECTERDocumentlevelRepresentation2020} and SciRepEval \cite{singhSciRepEvalMultiFormatBenchmark2023}.
Relevant to this work, OpenMSD \cite{Gao2023OpenMSDTM} presents a benchmark for scientific literature which contains text from papers spanning 103 languages. Notably, this is the first benchmark for scientific documents covering multiple languages.





\begin{figure*}[t]
    \centering
    \includegraphics[width=\linewidth]{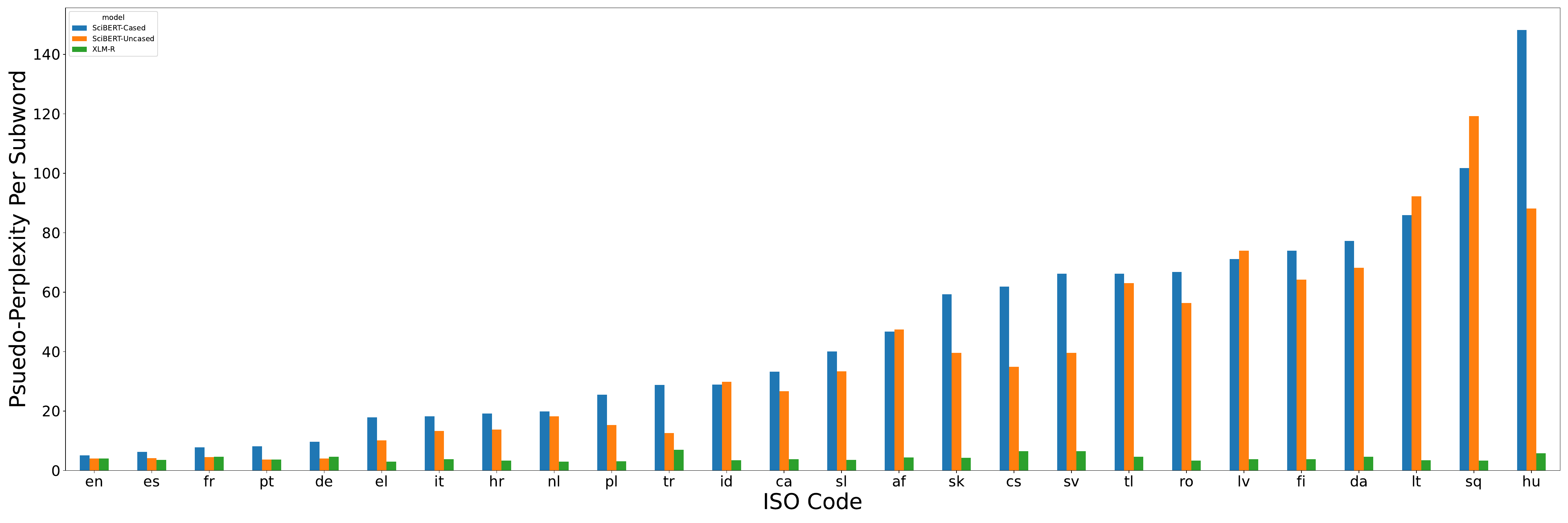}
    \caption{Pseudo-Perplexity (PPPL) per subword for languages with low unknown token counts. }
    \label{fig:perpl_fig}
\end{figure*}

\begin{figure}[t]
    \centering
    \includegraphics[width=\linewidth]{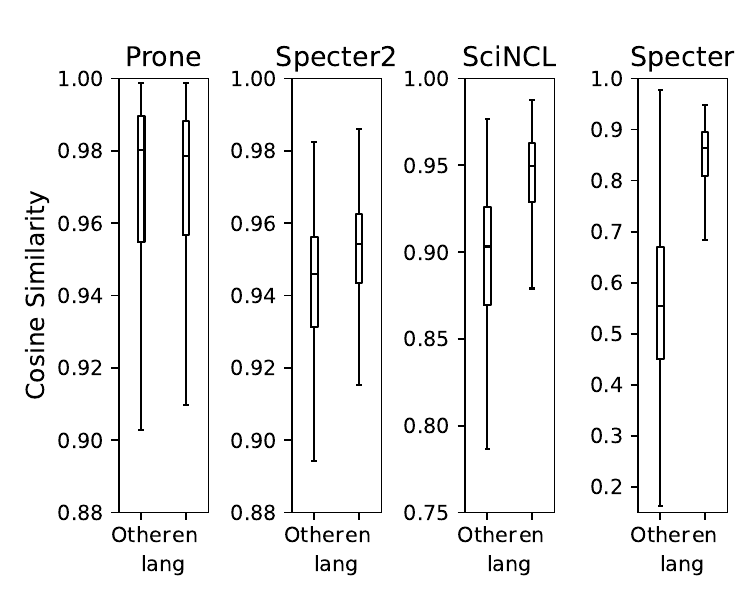}
    \caption{Relatively small cosines for non-English papers suggest opportunities for improvement.
}
    \label{fig:centroid_assumption_main}
\end{figure}


\section{How multilingual is the literature?}

To determine the multilinguality of the scientific literature, 
we take three samples of abstracts from a bulk download of Semantic Scholar\footnote{\url{https://www.semanticscholar.org/product/api}} from June 20, 2023. 
We use \texttt{langdetect}\footnote{\url{https://pypi.org/project/langdetect/}} and \texttt{langid.py}\footnote{\url{https://pypi.org/project/langid/}} to classify each abstract, and only keep papers where the two models agree. 
Across the three sets, the average sample size is 30.19m abstracts. Filtering abstracts which are too short for \texttt{langdetect}, or do not meet a minimum language confidence of 0.95, leaves 29.4m abstracts. Taking the intersection of the two detection algorithms, the final average sample size is 28.58m with a standard deviation of 2,385. 

We present results in \cref{fig:main_fig}. While English is clearly dominant, taking a 85.11\% share, we detect a surprisingly large number of other languages -- 50 total -- in the remaining abstracts. We list the languages along with their counts 
in the Appendix.  


On the larger end, papers in 
French, Spanish, Chinese, Portuguese, and Indonesian make up 10.37\% of the samples on average. Of these, the 3rd largest is Chinese, which while representing a share of over 600k papers, the vast majority of input tokens, 92.95\% and 93.14\% on average for cased and uncased SciBERT respectively, are represented as \texttt{UNK}, effectively leading to random outputs. For comparison, the average number of unknown characters for a multilingual model, XLM-R \cite{conneauUnsupervisedCrosslingualRepresentation2020}, is 0.06\%. 
While the remaining 45 languages cumulatively make up less than 5\% of the sample,
they cannot be ignored as the scale of the scientific literature must be taken into account; these languages still account for 1.32m papers. 

Although a considerable portion of the Semantic Scholar database consists of papers written in a language other than English. In contrast, prevalent papers written about scientific document representation, either models or datasets, are solely focused on English. More worryingly, is that these papers also fail the Bender Rule \cite{benderBenderRuleNamingLanguages2019}: not only do they not cover other languages, but they do not mention what languages their experiments or data cover. The use of English is implied and taken as the default language of research papers. 

This lack of evaluation datasets with 
non-English examples 
creates a blind spot in our understanding of how models perform on these papers, and causes a feedback loop: without an existing dataset, researchers will generally not be motivated to train models on other languages, for two reasons. 
First, there will be no straightforward way to know how well the model works for other languages, and second, for a fixed model size, introducing other languages dilutes the English representation of the model, which may lower its performance on existing benchmarks, even though the model performs better in a real-world setting. The introduction of OpenMSD represents a strong step in the right direction, although the creation and release of additional benchmarks is required for more comprehensive evaluation.
 
\section{Implicit Multilingual Evaluation}

\subsection{Unknown Scripts}
\label{sec:unk_scripts}

Models 
trained with an English-only vocabulary 
represent abstracts written in non-Latin scripts with many unknown tokens. If too many input tokens are unknown, 
the model will solely rely on parts of the text which are written in the supported scripts, 
often numbers or named entities. This can lead to erroneous embeddings which represent the document based on a few 
key words rather than the complete abstract. Even if the abstract includes an English translation at the end, the \texttt{UNK} tokens from the source script at the beginning may force the English portions outside the maximum supported sequence length.

To quantify the impact of using only a Latin-script vocabulary, we calculate the average percentage of \texttt{UNK} tokens contained in the tokenized abstracts for each language in our sample. \cref{tab:main_results} in the Appendix shows that for languages such as Chinese, Korean, Japanese, Arabic, and Hindi, more than 60\% of the input tokens are \texttt{UNK}. Overall, 14 total languages, which make up 973k abstracts or 3.3\% of our sample, fall under this constraint. Compared to the tokenization of a multilingual model, the maximum percentage of unknown tokens for any given language is 2.98\%.

\subsection{Perplexity on Seen Scripts}

To intrinsically evaluate performance on languages without a large number of \texttt{UNK} tokens, we calculate a pseudo-perplexity (PPPL), or model score \cite{salazarMaskedLanguageModel2020}. 
We present results in \cref{tab:perp_results} (in Appendix) and \cref{fig:perpl_fig}. While for some languages the PPPL is low, indicating that the model may be capable of producing reasonable outputs, it rapidly increases as we consider other languages, indicating deteriorating performance. For comparison, the PPPL of XLM-R remains consistently low across all languages. This motivates the use of a vocabulary which covers multiple scripts. Even if the perplexity after pretraining is high, it still allows for finetuning on task-specific data without any architectural changes. 

\subsection{Dependencies on Language \& Embedding}
\label{sec:emb_eval}


This section will use the centroid assumption
\[vec(v) \approx \sum_{v_j \in ref(v)} vec(v_j)\]
to test for possible issues with non-English languages.
This assumption approximates a document vector as the sum of its references.

The boxplots in Figure~\ref{fig:centroid_assumption_main} use cosines to compare documents vectors with the centroids of their references.
The differences between English and other languages
are relatively small for ProNE \cite{zhangProNEFastScalable2019}, a graph-based method.
In contrast, of the text-based models, Specter clearly produces worse representations for non-English papers. Other methods, SciNCL \cite{ostendorff-etal-2022-neighborhood} and Specter2, show clear improvements, but still do not reach the same performance of ProNE. Our hope is that for a truly multilingual model, the quality of representations for all languages will approach that of the graph-based methods.

\section{Real World Implications}

To highlight the real-world implications of deploying English-only models for a multilingual domain, we lookup papers in our sample to see how they are displayed on the Semantic Scholar website. We focus on the "tl;dr" section which, at the time of writing, is based on an English only generation model \cite{cacholaTLDRExtremeSummarization2020}. We sample papers written in Spanish, French, Japanese, Korean, and Farsi, and present a hand-picked selection in \cref{tab:real_world_tab}\footnote{Samples are picked to highlight cases which both work well and fail. The full sample will be released.}. Extractions for Spanish and French papers are reasonable -- outputs are often directly copied from the abstract but do not present any other information. For languages with non-Latin script, the performance is random and nonsensical at best, and potentially biased at worse as in the case of Example 3. 




\section{Future Directions}

Here, we discuss possible future directions to address the current lack of multilingual support for scientific document representation. 

\paragraph{Language Detection and Translation} English-only models are often used directly to provide end-users with recommendations or other suggestions, even though the query paper may be in an unsupported language. 
A simple option is to detect the abstract languages and either not generate outputs, or warn the user that the results may be unreliable for unsupported languages. For abstracts which have a translation, the English portion can be detected and used as input. Machine translation systems can also be used to translate abstracts; while this approach will introduce noise, we believe the outputs will be more reasonable, particularly for non-Latin script languages. 




\paragraph{Using Text Agnostic Models} As shown in \cref{sec:emb_eval}, graph-based approaches are less sensitive to language. 
As these methods will not be directly applicable for papers with no links, a combination of text-based, and graph-based approaches will likely result in the best performance. 

\paragraph{Training Multilingual Models} This solution 
is the simplest solution which offers the largest potential improvement. 
Multilingual models can leverage zero-shot transfer to make a better than random prediction, even if the language is not heavily trained on.
At the very least, including a vocabulary which covers other scripts would correct the unknown token problem, 
 and open the door for 
 other users to 
 continue the model pretraining on 
 other languages. 

\paragraph{Evaluation Datasets in Other Languages} The direction which we believe is most important, is to create multilingual benchmark datasets, for three reasons. 
First, benchmark datasets outlive specific models. A sufficiently difficult benchmark can be used by multiple iterations of models. Second, the creation of benchmark dataset can motivate others to create multilingual models. By creating a readily available way to evaluate the multilingual ability of models, this may motivate teams which would otherwise choose to create an English-only model to instead train a multilingual one. Finally, a benchmark datasets would allow us to evaluate existing pretrained models on their multilingual ability. As an example,  multilingual large language models have been shown to have strong zero-shot performance. The creation of a multilingual benchmark would allow us to evaluate these models on scientific document tasks, potentially revealing strong performance without requiring additional training. 

\section{Conclusion} 
In this work, we have highlighted the linguistic diversity found in the scientific literature, and shown that using English-only models in this domain leads to poor support for other languages. As the outputs of these models are often directly user-facing, we hope that this work motivates the community to create new models and benchmarks which can support document representation for all languages. 

\section{Ethics}

We do not believe there are any ethical concerns which arise from our work. We rely on commonly used and public APIs, models, and datasets.

\section{Limitations}

Many of the evaluations we present in our work are \textit{intrinsic} and while we assume that they are correlated with downstream performance on an \textit{extrinsic} task, such as paper recommendation, further experiments are necessary to confirm this hypothesis. Furthermore, while we have taken steps to reduce potential noise in our language detection, by using two algorithms, only keeping papers on which they both agree, and only keeping papers with a large confidence (>0.95), there is still likely to be some erroneous detections. We have manually reviewed a sample of the language classifications, and removed languages which were consistently erroneously. As such we believe our general findings remain valid, regardless of language detection noise present. 

\section*{Acknowledgments} The majority of this work was completed in the 2023 Jelinek Memorial Summer Workshop on Speech and Language Technologies (JSALT) and was supported with funds from Johns Hopkins University and ESPERANTO, the EU Horizons 2020 program’s Marie Sklodowska-Curie Grant No 101007666. The authors would like to thank members of the Better Together team, particularly Peter Vickers and Shabnam Tafreshi for their help, insights, and feedback on this work. 

\bibliography{bibs/zotero_bibtex,bibs/kwc}
\clearpage

\appendix
\onecolumn
\section{Appendix}
\label{sec:appendix}

\begin{table*}[h]
    \centering
    \tiny
    \begin{adjustbox}{width=\textwidth}
\begin{tabular}{@{\extracolsep{4pt}}lllrrrrrrr@{}}
\toprule
      &              &            & \multicolumn{4}{c}{Language Counts}          & \multicolumn{3}{c}{Mean Prop. Unk}        \\
      \cline{4-7} \cline{8-10}\\
\texttt{ISO}   & \texttt{Language}     & \texttt{Script}     & \texttt{Mean}    & \texttt{STD}   & \texttt{Mean Prop.} & \texttt{NE Mean Prop.} & \texttt{SciBERT-C} & \texttt{SciBERT-U} & \texttt{XLM-R}   \\
\midrule
en    & English      & Latin      & 24,862k & 2.20k & 85.11\%    &    -           & 00.08\%       & 00.08\%         & 00.00\% \\
fr    & French       & Latin      & 770k    & .75k  & 2.64\%     & 17.70\%       & 00.08\%       & 00.08\%         & 00.00\% \\
es    & Spanish      & Latin      & 661k    & .55k  & 2.26\%     & 15.20\%       & 00.02\%       & 00.02\%         & 00.00\% \\
zh & \textbf{Chinese}      & Chinese    & 633k    & 1k    & 2.17\%     & 14.55\%       & 92.95\%       & 93.14\%         & 00.06\% \\
pt    & Portuguese   & Latin      & 488k    & .52k  & 1.67\%     & 11.22\%       & 00.02\%       & 00.02\%         & 00.00\% \\
id    & Indonesian   & Latin      & 478k    & .26k  & 1.64\%     & 11.00\%       & 00.03\%       & 00.02\%         & 00.00\% \\
de    & German       & Latin      & 283k    & .40k  & 0.97\%     & 6.50\%        & 00.04\%       & 00.04\%         & 00.00\% \\
ko    & \textbf{Korean}       & Korean     & 227k    & .44k  & 0.78\%     & 5.22\%        & 81.66\%       & 81.97\%         & 00.11\% \\
ru    & Russian      & Cyrillic   & 112k    & .47k  & 0.38\%     & 2.57\%        & 10.81\%       & 10.07\%         & 00.00\% \\
tr    & Turkish      & Latin      & 94k     & .08k  & 0.32\%     & 2.15\%        & 00.07\%       & 00.07\%         & 00.00\% \\
ja    & \textbf{Japanese}     & Japanese   & 89k     & .24k  & 0.30\%     & 2.04\%        & 86.00\%       & 82.01\%         & 00.03\% \\
fa    & Persian      & Arabic     & 66k     & .23k  & 0.22\%     & 1.51\%        & 49.71\%       & 24.63\%         & 00.00\% \\
sv    & Swedish      & Latin      & 57k     & .16k  & 0.20\%     & 1.32\%        & 00.01\%       & 00.01\%         & 00.00\% \\
cs    & Czech        & Latin      & 57k     & .14k  & 0.20\%     & 1.32\%        & 00.03\%       & 00.03\%         & 00.00\% \\
hr    & Croatian     & Latin      & 48k     & .06k  & 0.16\%     & 1.11\%        & 00.02\%       & 00.02\%         & 00.00\% \\
it    & Italian      & Latin      & 46k     & .13k  & 0.16\%     & 1.07\%        & 00.06\%       & 00.06\%         & 00.00\% \\
uk    & Ukrainian    & Cyrillic   & 40k     & .21k  & 0.14\%     & 0.92\%        & 15.54\%       & 14.16\%         & 00.00\% \\
nl    & Dutch        & Latin      & 39k     & .03k  & 0.13\%     & 0.90\%        & 00.02\%       & 00.01\%         & 00.00\% \\
pl    & Polish       & Latin      & 36k     & .06k  & 0.12\%     & 0.83\%        & 00.10\%       & 00.10\%         & 00.00\% \\
ar    & \textbf{Arabic}       & Arabic     & 24k     & .10k  & 0.08\%     & 0.54\%        & 61.97\%       & 15.78\%         & 00.04\% \\
ca    & Catalan      & Latin      & 23k     & .12k  & 0.08\%     & 0.54\%        & 00.41\%       & 00.36\%         & 00.00\% \\
hu    & Hungarian    & Latin      & 10k     & .05k  & 0.03\%     & 0.23\%        & 00.03\%       & 00.03\%         & 00.00\% \\
sl    & Slovenian    & Latin      & 10k     & .09k  & 0.03\%     & 0.23\%        & 00.02\%       & 00.02\%         & 00.00\% \\
el    & Modern Greek & Greek      & 9k      & .06k  & 0.03\%     & 0.22\%        & 00.75\%       & 00.75\%         & 00.01\% \\
vi    & Vietnamese   & Latin      & 8k      & .05k  & 0.03\%     & 0.19\%        & 18.08\%       & 18.77\%         & 00.02\% \\
no    & Norwegian    & Latin      & 6k      & .05k  & 0.02\%     & 0.14\%        & 04.99\%       & 05.11\%         & 00.01\% \\
fi    & Finnish      & Latin      & 5k      & .09k  & 0.02\%     & 0.11\%        & 00.01\%       & 00.02\%         & 00.00\% \\
ro    & Romanian     & Latin      & 4k      & .10k  & 0.02\%     & 0.10\%        & 00.52\%       & 00.45\%         & 00.01\% \\
da    & Danish       & Latin      & 4k      & .09k  & 0.02\%     & 0.10\%        & 00.11\%       & 00.09\%         & 00.00\% \\
af    & Afrikaans    & Latin      & 4k      & .06k  & 0.01\%     & 0.10\%        & 00.06\%       & 00.05\%         & 00.00\% \\
th    & Thai         & Thai       & 4k      & .03k  & 0.01\%     & 0.09\%        & 49.57\%       & 50.06\%         & 00.02\% \\
lt    & Lithuanian   & Latin      & 4k      & .06k  & 0.01\%     & 0.08\%        & 00.18\%       & 00.18\%         & 00.01\% \\
sk    & Slovak       & Latin      & 3k      & .01k  & 0.01\%     & 0.07\%        & 00.07\%       & 00.08\%         & 00.01\% \\
et    & Estonian     & Latin      & 1k      & .01k  & 0.00\%     & 0.03\%        & 04.67\%       & 04.69\%         & 00.00\% \\
bg    & Bulgarian    & Cyrillic   & 1k      & .02k  & 0.00\%     & 0.03\%        & 17.34\%       & 14.88\%         & 00.00\% \\
lv    & Latvian      & Latin      & 1k      & .02k  & 0.00\%     & 0.02\%        & 00.01\%       & 00.01\%         & 00.00\% \\
mk    & Macedonian   & Cyrillic   & 1k      & .02k  & 0.00\%     & 0.02\%        & 22.10\%       & 19.67\%         & 00.01\% \\
hi    & \textbf{Hindi}        & Devanagari & 372     & 11    & 0.00\%     & 0.01\%        & 92.59\%       & 92.62\%         & 00.13\% \\
sw    & Swahili      & Latin      & 345     & 18    & 0.00\%     & 0.01\%        & 00.82\%       & 00.86\%         & 00.00\% \\
he    & \textbf{Hebrew}       & Hebrew     & 282     & 24    & 0.00\%     & 0.01\%        & 72.35\%       & 72.63\%         & 00.32\% \\
ne    & \textbf{Nepali}       & Devanagari & 181     & 7     & 0.00\%     & 0.00\%        & 93.20\%       & 93.24\%         & 00.00\% \\
tl    & Tagalog      & Latin      & 162     & 4     & 0.00\%     & 0.00\%        & 00.31\%       & 00.30\%         & 00.00\% \\
sq    & Albanian     & Latin      & 151     & 7     & 0.00\%     & 0.00\%        & 00.03\%       & 00.03\%         & 00.00\% \\
mr    & \textbf{Marathi}      & Devanagari & 53      & 8     & 0.00\%     & 0.00\%        & 87.02\%       & 87.06\%         & 00.00\% \\
gu    & \textbf{Gujarati}     & Gujarati   & 38      & 8     & 0.00\%     & 0.00\%        & 87.85\%       & 87.88\%         & 00.85\% \\
ta    & \textbf{Tamil}        & Tamil      & 29      & 3     & 0.00\%     & 0.00\%        & 66.09\%       & 66.39\%         & 00.00\% \\
ml    & \textbf{Malayalam}    & Malayalam  & 13      & 2     & 0.00\%     & 0.00\%        & 77.29\%       & 77.53\%         & 02.98\% \\
ur    & \textbf{Urdu}         & Arabic     & 13      & 3     & 0.00\%     & 0.00\%        & 60.57\%       & 42.69\%         & 00.78\% \\
bn    & Bengali      & Latin      & 10      & 2     & 0.00\%     & 0.00\%        & 06.96\%       & 06.91\%         & 02.38\% \\
pa    & \textbf{Panjabi}      & Latin      & 7       & 3     & 0.00\%     & 0.00\%        & 68.58\%       & 69.30\%         & 00.00\% \\
kn    & \textbf{Kannada}      & Kannada    & 4       & 3     & 0.00\%     & 0.00\%        & 82.67\%       & 82.79\%         & 00.03\% \\
\bottomrule
\end{tabular}
\end{adjustbox}
    \caption{ \label{tab:main_results}Main Results}
   
\end{table*}

\begin{table}[]
\begin{tabular}{@{}llrrrr@{}}
\toprule
ISO & Language             & SciBERT-Cased & SciBERT-Uncased & XLM-R & Sample Size \\
\midrule
en  & English              & 05.07         & 04.04           & 04.05 & 4985        \\
es  & Spanish              & 06.26         & 04.15           & 03.55 & 4988        \\
fr  & French               & 07.86         & 04.49           & 04.65 & 4985        \\
pt  & Portuguese           & 08.22         & 03.74           & 03.70 & 4989        \\
de  & German               & 09.69         & 04.08           & 04.67 & 4965        \\
el  & Greek & 17.92         & 10.10           & 02.99 & 4984        \\
it  & Italian              & 18.26         & 13.34           & 03.83 & 4982        \\
hr  & Croatian             & 19.19         & 13.83           & 03.40 & 4981        \\
nl  & Dutch                & 19.89         & 18.30           & 02.95 & 4985        \\
pl  & Polish               & 25.52         & 15.37           & 03.12 & 4994        \\
tr  & Turkish              & 28.85         & 12.59           & 06.94 & 4995        \\
id  & Indonesian           & 28.92         & 29.89           & 03.43 & 4996        \\
ca  & Catalan              & 33.30         & 26.67           & 03.84 & 4906        \\
sl  & Slovenian            & 40.05         & 33.40           & 03.54 & 4852        \\
af  & Afrikaans            & 46.78         & 47.48           & 04.35 & 2576        \\
sk  & Slovak               & 59.25         & 39.61           & 04.31 & 2149        \\
cs  & Czech                & 61.92         & 34.87           & 06.46 & 4998        \\
sv  & Swedish              & 66.19         & 39.59           & 06.55 & 4970        \\
tl  & Tagalog              & 66.23         & 63.06           & 04.68 & 158         \\
ro  & Romanian             & 66.86         & 56.35           & 03.36 & 1837        \\
lv  & Latvian              & 71.15         & 73.97           & 03.86 & 611         \\
fi  & Finnish              & 73.92         & 64.27           & 03.82 & 3110        \\
da  & Danish               & 77.21         & 68.22           & 04.59 & 2946        \\
lt  & Lithuanian           & 85.94         & 92.23           & 03.43 & 3338        \\
sq  & Albanian             & 101.74        & 119.27          & 03.35 & 133         \\
hu  & Hungarian            & 148.25        & 88.12           & 05.87 & 4925        \\
\bottomrule
\end{tabular}
    \caption{Perplexity Results}
    \label{tab:perp_results}
\end{table}
\onecolumn
\renewcommand{\arraystretch}{2}
\begin{xltabular}{\textwidth}{lllXX}

\toprule
Num. & ISO & CorpusID  & Title   & TL;DR    \\
\bottomrule
\endfirsthead
\toprule
Num. & ISO & CorpusID  & Title   & TL;DR    \\
\bottomrule
\endhead
\bottomrule
&&&\multicolumn{2}{c}{\textit{Continued on next page}}  \\
\bottomrule
\endfoot
\bottomrule
\caption{Hand-selected examples which highlight errors in the TL;DR model for papers written in Non-English, Non-Latin scripts. TL;DRs are collected using the Semantic Scholar API on 11/29/23.} \label{tab:real_world_tab}
\endlastfoot
1    & es  & \href{https://www.semanticscholar.org/paper/c45f6790905f3800c7197d7f0862c8e1655f151c}{90472702}  & Integración productiva ganadero-forestal: evaluación de impacto en dos clubes agrarios en el departamento de Durazno, Uruguay                 & El trabajo tuvo como objetivo evaluar el impacto socioeconomico sobre dos clubes agrarios del Movimiento de Juventud Agraria del Acuerdo de Cooperacion with la empresa forestal Montes del Plata en el marco del Programa de Integracion Productiva de la misma, en el departamento de Durazno, Uruguay.     \\
2    & es  & \href{https://www.semanticscholar.org/paper/57ba552077daada55878126e5f211364c3bb27db}{71074825}  & El manejo multidisciplinario de la lesión de vía biliar                                                                                       & Las fugas por lesiones laterales son susceptibles de tratarse con maniobras that disminuyen the presion ductal (esfinterotomia mas endoprotesis), y mientras exista continuidad de la via biliar, el tratamiento endoscopico esta indicado.                                                                   \\
3    & fa  & \href{https://www.semanticscholar.org/paper/3311a0b0842a37bd7ad7aa4c7d84c777183032c3}{249368552} & Effect Of Dimensions Smart Attached Mass And Local Stiffness On Dynamic Response Of Thick Plates                                              & It is revealed that the number of people affected by the recent Paris attacks has increased by more than 50\% in the past year, with the total number of fatalities increasing from 11 to more than 200.                                                                                                      \\
4    & fa  & \href{https://www.semanticscholar.org/paper/31d0bd62d2ad3c22087242872209617160c5f7dc}{166356916} & A comparison between Iran and the Selected Countries on the Efficiency of expenses in Health Sector by Data Envelopment Analysis (1998-2008). & The DEA has confirmed that it will not be pursuing legal action over the use of the word “toxic” in connection with the death of former DEA agent David Bailey.                                                                                                                                               \\
5    & fr  & \href{https://www.semanticscholar.org/paper/eb72956fe91c8e6ee00047fffd5bb2376faac1c7}{71021028}  & Les hépatites et les atteintes des voies biliaires dues à l'amoxicilline-acide clavulanique                                                   & L'evolution est habituellement favorable and the guerison complete s'observe en quelques semaines, mais plusieurs arguments sont en faveur d'une origine immuno-allergique.                                                                                                                                   \\
6    & fr  & \href{https://www.semanticscholar.org/paper/7636b42cf5ef64b97b18a127a563917f852e642f}{90599195}  & Sensibilité des races ovines et caprines tchadiennes au virus de la peste des petits ruminants                                                & Il a ete egalement constate that les moutons de race Sahel paraissent plus sensibles que les chevres deRace Sahel.                                                                                                                                                                                            \\
7    & fr  & \href{https://www.semanticscholar.org/paper/ffd9f96d01d58653316e041c86f4a1c482b289b9}{216003344} & Putting on a show: using computer graphics to train end-users                                                                                 & Les bibliotheques de l'Universite de Pennsylvanie ont mis au point une formule basee sur un projecteur video connecte a l'ecran d'un ordinateur personnel, pour former leurs utilisateurs a la recherche d'informations en ligne.                                                                             \\
8    & ja  & \href{https://www.semanticscholar.org/paper/010b60c43656737c90a3e942815adf895a283709}{76251764}  & Clinical evaluation of Tranilast for keloid and hyperplasic scar.Double blind comparative test using heparinoid ointment as the control drug. & It’s time to dust off the sledgehammers and start cleaning up after yourselves.                                                                                                                                                                                                                               \\
9    & ja  & \href{https://www.semanticscholar.org/paper/eef307904967645830059ab3508d95419a14c494}{31383396}  & {[}Protective efficacy of antiserum against Bacteroides gingivalis{]}.                                                                        & It is confirmed that Bacteroides gingivalis is a candidate for the Nobel Prize in physiology or medicine.                                                                                                                                                                                                     \\
10   & ja  & \href{https://www.semanticscholar.org/paper/9580d7995d8436470ccf69809c1e3e6481049037}{87290487}  & Mortality Factors of the Pine-moth, Dendrolimus spectabilis BUTLER in the Suburbs of Chiba City                                               & It is revealed that, for the first time, a non-European country, the U.S. government has acknowledged homosexuality.                                                                                                                                                                                          \\
11   & ja  & \href{https://www.semanticscholar.org/paper/dc738baab9110bc2497399413021e3e1730e1973}{71808165}  & INTESTINAL OBSTRUCTION CAUSED BY SEVERE OMENTO-MESENTERIC PANNICULITIS FOLLOWING RECTAL CANCER SURGERY                                        & It’s time to get ready for winter, people!                                                                                                                                                                                                                                                                    \\
12   & ja  & \href{https://www.semanticscholar.org/paper/ff0adc99fad8f8696e8cc22444d22d9a3b2ddd7d}{84854659}  & Existence of a stem-cell lineage in an infectious venereal tumor of the dog                                                                   & The history of stem cell research, Â£1.5bn in research and development, and more.                                                                                                                                                                                                                             \\
13   & ja  & \href{https://www.semanticscholar.org/paper/d831655c4c5246fbafe447f7c36096a9ec084512}{84779794}  & Relation between ants and aphids in a citrus orchard.                                                                                         & Five-year-old boy with Down's syndrome is diagnosed with encephalitis.                                                                                                                                                                                                                                        \\
14   & ja  & \href{https://www.semanticscholar.org/paper/7508473cc518ce306ef2d6973a189d60c2c223b2}{57713022}  & Lung Cancer Mass Screening with Sputum Cytology in Japan                                                                                      & The results of famous randomized control studies conducted in 1970s and 1980s in the US and those of Miyagi Prefecture in 1990s showed that higher sensitivity for lung cancer detection and better prognosis in the 1990s, suggesting that lung cancer screening with sputum cytology is important in Japan. \\
15   & ko  & \href{https://www.semanticscholar.org/paper/e5d9bfe2dd5d57cb0ec4167b4ba609c810e554be}{203819581} & PPG lul i-yong-han sim-hyel-kwan cil-hwan yey-chuk si-su-theym-uy sel-kyey mich kwu-hyen                                                                                                               & The analyzed results which show the heart rate variability and the distribution of heart rate for before and after using PPG are proposed.         \\
\end{xltabular}
\clearpage



\end{document}